\documentclass[conference,a4paper]{IEEEtran} 
\IEEEoverridecommandlockouts
\usepackage{cite}
\usepackage{amsmath,amssymb,amsfonts}
\usepackage{graphicx}
\usepackage{textcomp}
\usepackage{xcolor}
\usepackage{adjustbox}
\usepackage{algorithm}
\usepackage[noend]{algpseudocode}
\usepackage{chngpage}
\usepackage{subfig}
\def\BibTeX{{\rm B\kern-.05em{\sc i\kern-.025em b}\kern-.08em
    T\kern-.1667em\lower.7ex\hbox{E}\kern-.125emX}}
    
\makeatletter 
\newcommand{\linebreakand}{%
  \end{@IEEEauthorhalign}
  \hfill\mbox{}\par
  \mbox{}\hfill\begin{@IEEEauthorhalign}
}
\makeatother 
    
\begin{document}

\title{Robust Knowledge Adaptation for Federated Unsupervised Person ReID\\

\thanks{* Corresponding author.}
}

\author{\IEEEauthorblockN{Jianfeng Weng}
\IEEEauthorblockA{\textit{School of Computer Science} \\
\textit{The University of Sydney}\\
Sydney, Australia \\
jwen0609@uni.sydney.edu.au}
\and
\IEEEauthorblockN{Kun Hu\textsuperscript{*}}
\IEEEauthorblockA{\textit{School of Computer Science} \\
\textit{The University of Sydney}\\
Sydney, Australia \\
hukun\_sdu@hotmail.com}
\and
\IEEEauthorblockN{Tingting Yao}
\IEEEauthorblockA{\textit{Department of Electronic Information Engineering} \\
\textit{Dalian Maritime University}\\
Dalian, China \\
ytt1030@dlmu.edu.cn}
\and
\linebreakand
\IEEEauthorblockN{Jingya Wang}
\IEEEauthorblockA{\textit{School of Information Science and Technology} \\
\textit{ShanghaiTech University}\\
Shanghai, China \\
wangjingya@shanghaitech.edu.cn}
\and
\IEEEauthorblockN{Zhiyong Wang}
\IEEEauthorblockA{\textit{School of Computer Science} \\
\textit{The University of Sydney}\\
Sydney, Australia \\
zhiyong.wang@sydney.edu.au}
}

\maketitle

\begin{abstract}
Person Re-identification (ReID) has been extensively studied in recent years due to the increasing demand in public security. However, collecting and dealing with sensitive personal data raises privacy concerns. 
Therefore, federated learning has been explored for Person ReID, which aims to share minimal sensitive data between different parties (clients). 
However, existing federated learning based person ReID methods generally rely on laborious and time-consuming data annotations and it is difficult to guarantee cross-domain consistency. 
Thus, in this work, a federated unsupervised cluster-contrastive (FedUCC) learning method is proposed for Person ReID. FedUCC introduces a three-stage modelling strategy following a coarse-to-fine manner. In detail, generic knowledge, specialized knowledge and patch knowledge are discovered using a deep neural network. This enables the sharing of mutual knowledge among clients while retaining local domain-specific knowledge based on the kinds of network layers and their parameters. 
Comprehensive experiments on 8 public benchmark datasets demonstrate the state-of-the-art performance of our proposed method. 
\end{abstract}


\section{Introduction}
\label{sec:intro}

Person re-identification (ReID) aims to identify a given person across non-overlapped camera views and has been widely studied due to its great potential for the public security and forensics
\cite{qiao2018few}. 
Over the years, various types of methods have been studied, including supervised
learning and unsupervised learning. 
Recently, cross-domain modelling becomes a new direction in the field of person ReID, which exploits the diversity of multi-domain datasets to enhance generalization capability~\cite{dai2021generalizable,jia2019frustratingly,jin2020style}.
Conventionally, the majority of multi-domain ReID solutions adopt a centralised mechanism that gathers all datasets to a single location for modelling. However, there have been increasing concerns on such centralised person ReID methods due to data privacy issues among sharing sensitive personal data across multiple domains or clients~\cite{custers2019eu}.

\begin{figure}[t]
\begin{center}
  \includegraphics[width=0.95\linewidth]{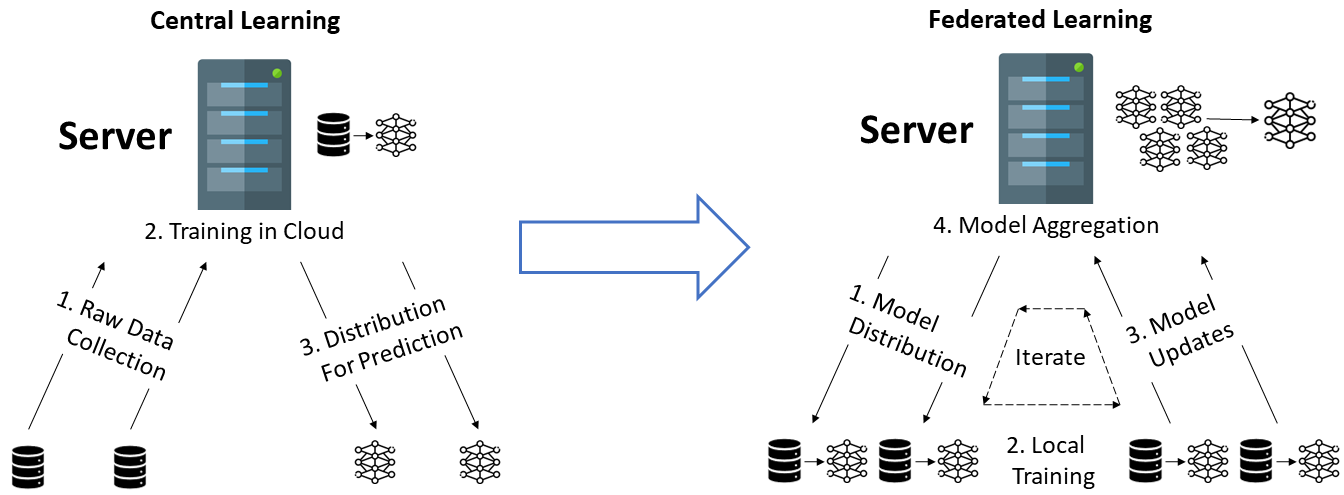}
\end{center}
  \caption{
  Illustrative comparison between centralised learning pipeline and decentralised pipeline of federated learning.
}
\label{fig:cmp_pipeline}
\end{figure}

Therefore, decentralised learning paradigm, specifically federated learning, has gained increasing attentions recently. It allows models to be trained collaboratively across multiple sources or clients without data centralisation~\cite{mcmahan2017communication}. 
Figure \ref{fig:cmp_pipeline} illustrates the comparison between centralised learning and federated learning.
FedAVG~\cite{mcmahan2017communication}, for example, aggregated the model parameters at a central server after a number of local epochs, and the new global model is re-distributed towards clients as new parameters for local optimization.
Using this simple approach, multiple clients can collaborate without explicit data sharing. Hence, federated learning has been widely utilised in various applications where data privacy is seriously concerned, such as facial expression recognition~\cite{shome2021fedaffect} and medical image segmentation~\cite{liu2021feddg}. 

For Person ReID, a number of recent studies have explored federated learning strategy~\cite{zhuang2021joint,zhuang2020performance,zhang2021federated,sun2021decentralised,wu2021decentralised}.
For example, FedReID~\cite{zhuang2020performance} adopted FedAvg to improve the generalizability of learned models via 9 multi-domain datasets. 
However, most existing methods typically assume that the dataset of each client is annotated, which can be time-consuming and laborious to obtain. This limits the scalability for large-scale real-world deployment.
FedUReID~\cite{zhuang2021joint} attempted to resolve this issue by introducing an unsupervised learning method for federated Person ReID. Specifically, hierarchical clustering was used to progressively merge neighbouring clusters during training.
Nonetheless, determining the stopping criteria for clustering on each client is challenging, as the number of images and identities often varies widely.
In addition, the variation between clients could be impacted by statistical heterogeneity, such as label skew, feature skew and concept shift~\cite{kairouz2021advances}. 
Existing studies often fine-tune the global model using public datasets~\cite{wu2021decentralised},
and utilise adversarial learning~\cite{zhang2021federated} to alleviate such issue. However, a unified model is still not sufficient to perform well in this scenario, especially when considering the nature of the extreme statistical heterogeneity in the person ReID task. Model personalization methods attempted to incorporate generic and specialized knowledge using different model parameters~\cite{arivazhagan2019federated,liang2020think,li2021fedbn,shen2022cd2}. 
However, these methods have not addressed the possible domain over-fitting issues yet, with the potential generic knowledge overwhelmed by potential local knowledge bias. 

To address the aforementioned challenges, a three-stage learning strategy in a coarse-to-fine manner is proposed for federated unsupervised learning based Person ReID. 
First, the task is modelled with generic knowledge by a conventional federated learning scheme based on DBSCAN~\cite{ester1996density} clustering. This stage aggregates and distributes all parameters from local clients. 
Second, specialized knowledge is explored in pursuit of client personalization. This is achieved by decoupling the client-specific knowledge from generic knowledge via parameter localization. 
Third, we further enhance fine-grained patterns by incorporating patch-level feature alignment across clients. Unlike holistic person images, which present potential large variations from changing poses and view-angles, patch-level information provides fine and consistent person ReID representations. Finally, comprehensive experiments and analyses have been conducted to demonstrate the state-of-the-art performance of our proposed FedUCC on 8 public Person ReID datasets.

In summary, the major contributions of this paper are three-fold as follows:
\begin{itemize}
  \item A novel unsupervised federated learning method is proposed for Person ReID with a coarse-to-fine strategy including three learning stages. 
  \item A novel strategy to explore client specific knowledge by incorporating patch-level representations. 
  \item Comprehensive experiments demonstrate the state-of-the-art performance of the proposed method on 8 benchmark datasets.
\end{itemize}

\section{Related Work}
\subsection{Unsupervised Person Re-identification}
Generally, unsupervised person ReID methods can be categorised into two streams: unsupervised domain adaptation (UDA)~\cite{wei2018person,chen2019instance,ge2020mutual} and fully
unsupervised methods~\cite{isobe2021towards,xuan2021intra}.

\textbf{Unsupervised Domain Adaptation for Person ReID} is considered as a semi-supervised learning approach. It involves knowledge adaptation between the datasets of two domains: a labeled source domain and an unlabeled target domain.
Given a source domain with labeled data, the knowledge from the source domain is exploited to supervise the learning on an unlabeled target domain. 
Existing studies have explored various transfer learning techniques. There have been attempts~\cite{wei2018person,chen2019instance,deng2018image} to minimize the domain gap by transferring the source image style to the target image style using a GAN~\cite{zheng2019joint}, while preserving the contents. 
Knowledge distillation (e.g.~\cite{ge2020mutual}) and meta-learning (e.g.~\cite{zhao2021learning}) have also been adopted. A prerequisite of these methods is that both the source domain and the target domain are accessible during the training stage, which does not meet the criteria for protecting sensitive private data.

\textbf{Fully Unsupervised Person ReID}
can be more challenging as there is no prior knowledge (i.e., label annotations), compared with the UDA methods. 
Clustering-based methods such as Density-Based Spatial Clustering of Applications with Noise (DBSCAN) and K-mean are popularly used to estimate image identity relationships~\cite{ji2020attention,lin2019bottom}, which consider all the images in a cluster as the same class for model learning.
However, the noise label on misclustered images can degrade the performance. Several approaches address this issue by aligning image feature representations towards cluster centroids using contrastive learning~\cite{wang2021camera,dai2021cluster}.  
Thus, the mislabeled images of the same identity are not directly optimized to drift apart in the feature space.
BUC (Bottom Up Clustering)~\cite{lin2019bottom} employs a hierarchical clustering approach to iteratively associate similar images, beginning with each image as a single cluster, and terminate until a predefined number of clusters are remained.
Besides clustering, multi-labels were also used to indicate the identity relationship among images~\cite{wang2020unsupervised}.
For unsupervised learning in the federated setting, FedUReID~\cite{zhuang2021joint} followed the BUC~\cite{lin2019bottom} strategy to iteratively associate neighbouring clusters, and designed the 
cluster stopping criteria according to the batch precision score.
However, the positive correlation between clustering accuracy and batch precision score cannot be guaranteed as batch precision scores are highly dependent on the distribution of a dataset. Thus, both under-clustering and over-clustering remain a potential issue.
\begin{figure*}[!tbp]
  \centering
  \subfloat[A global round contains local training, global aggregation and distribution.]{\includegraphics[width=0.49\textwidth]{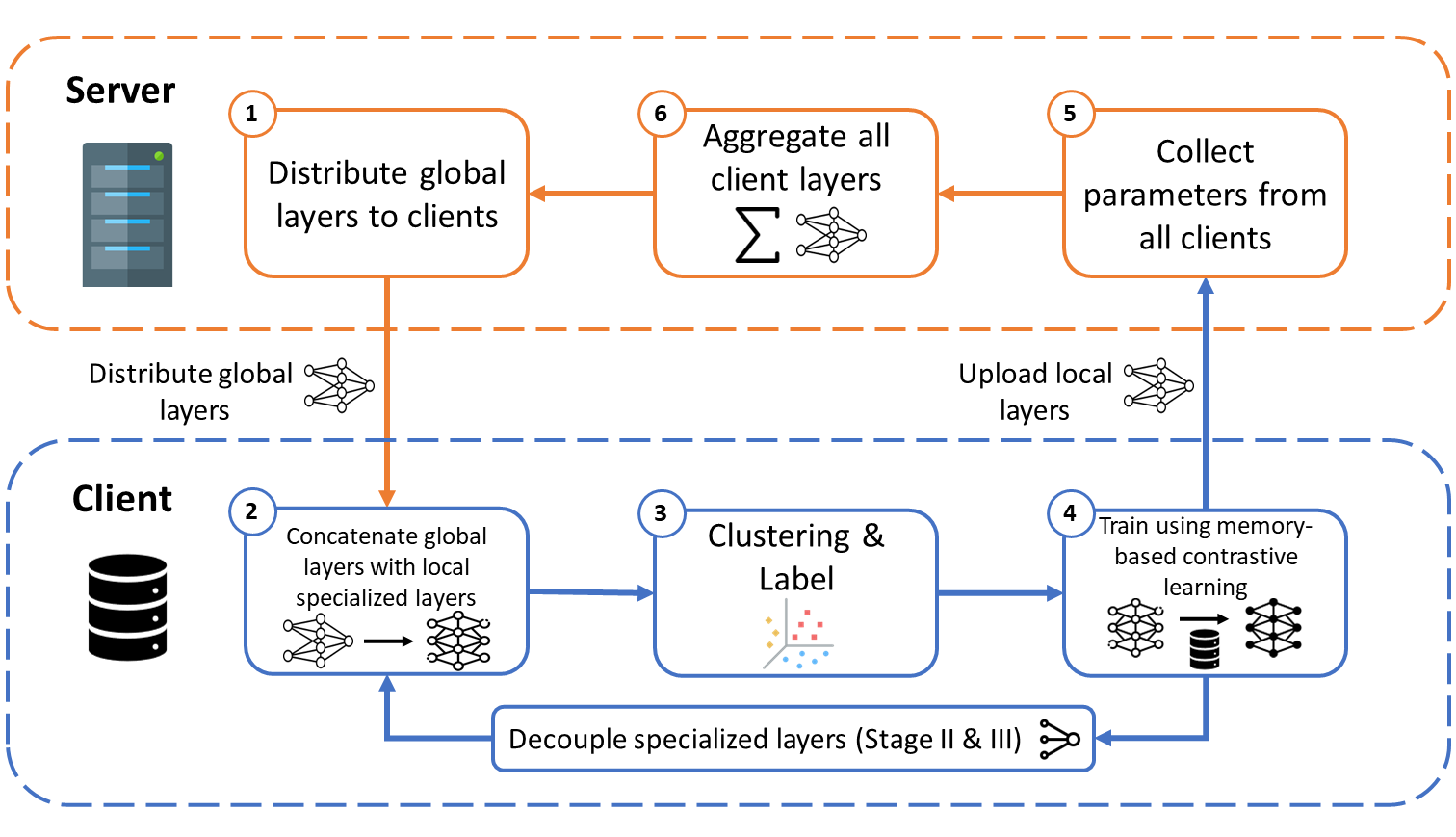}\label{fig:3stg_pipeline}}
  \hfill
  \subfloat[Three-stage local training.]{\includegraphics[width=0.49\textwidth]{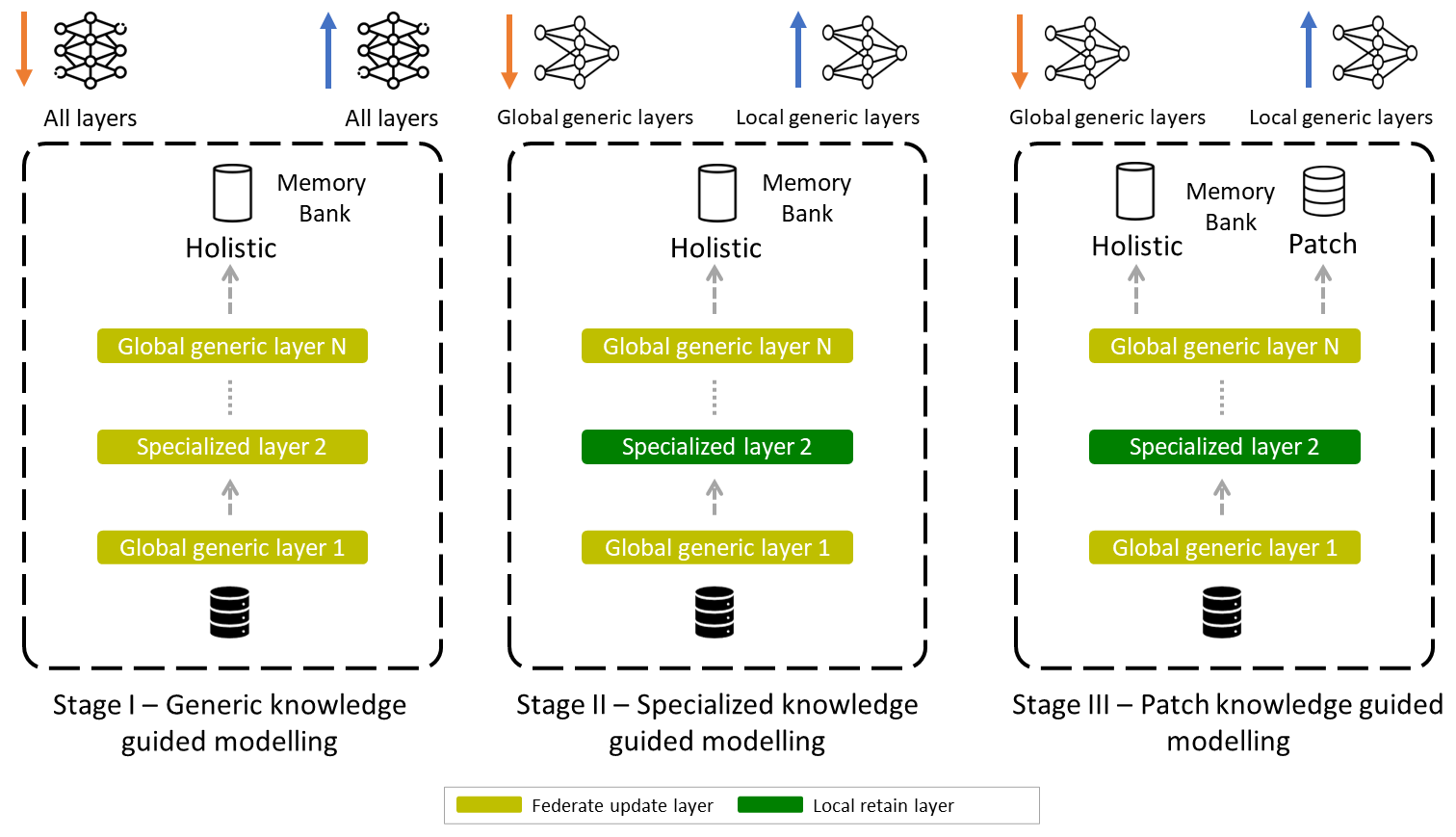}\label{fig:f2}}
  \caption{Illustration of the proposed FedUCC method: (a) a global round, and (b) three-stage strategy for local training.}
\end{figure*}
\subsection{Federated Learning}
Federated learning is proposed to address decentralized training and reduce the risk of privacy leakage. The intuition of the idea is to 
conduct knowledge exchange between the clients with abstract information rather than the raw inputs which can pose high privacy concerns. For example, trained model weights and gradients can be transferred for this purpose~\cite{mcmahan2017communication}. 

As federated learning is a relatively new research topic, the majority of existing studies focus on supervised learning~\cite{liu2021feddg, zhuang2020performance,zhang2021federated,shome2021fedaffect}. 
And among these studies, statistical heterogeneity is a core research problem. Several strategies have been proposed to alleviate its issue. 
A common strategy is to adopt extra regularization terms to ensure the model similarity of local models and averaged global model~\cite{li2020federated,li2021model}, using distance metric such as L2-normalization to regularize global model and local model distance~\cite{li2020federated} or image 
feature presentation similarity~\cite{li2021model}. While this can limit the impact of local updates and alleviate the client drift problem, it can restrict the model's ability to achieve maximum performance on all clients. 
Meta-learning techniques such as MAML(Model-Agnostic Meta Learning)~\cite{finn2017model} are used to obtain a good initial global model that can adapt quickly to a new heterogeneous task with few local gradient descent~\cite{fallah2020personalized}. However, this strategy is computational expensive because it requires the computation of second-order gradients.
Feddg~\cite{liu2021feddg} attempted to share local image distributions across clients by first decoupling style and content patterns using Fourier Transform~\cite{nussbaumer1981fast}. Then, by re-synthesizing images with these patterns for image style transfer, it allowed local models to train on augmented images with the distributions across clients. However, Fourier transform is only capable of decoding generic image color and texture, which is difficult to formulate the statistical heterogeneity between Person ReID datasets.

Compared with the abovementioned approaches which attempt to learn a strong global model that adapts to all clients, model personalization approach trains individual models for each client to focus on client-specific patterns~\cite{li2021fedbn,arivazhagan2019federated}. 
Instead of conducting model aggregation to exchange cross-client knowledge, knowledge consensus is obtained by gathering image predictions on a public dataset, and using knowledge distillation on averaged image predictions to conduct a knowledge exchange~\cite{li2019fedmd}.
Alternatively, parameter personalization approaches are proposed to store generic and specialized knowledge on separate parameters of the model, and the client-specific model is reformulated by integrating generic parameters and client-specific parameters.~\cite{li2021fedbn,arivazhagan2019federated,liang2020think,shen2022cd2}.
The selections of personal parameters were investigated in various works, including top layers~\cite{arivazhagan2019federated} and bottom layers~\cite{liang2020think} of the model network, as well as batch normalization~\cite{li2021fedbn} and convolutional channel parameters~\cite{shen2022cd2}, which are evenly distributed across the model network depth. 
These methods, however, fail to address potential domain over-fitting, in which local specialized knowledge overtakes generic knowledge, causing performance degrading for the more biased clients.
In this work, we address this by introducing a patch-level feature alignment with fine-feature consistency across clients. 

\section{Proposed Method}
\label{sec:method}

\subsection{Overview}
Fig.~\ref{fig:3stg_pipeline} illustrates the overview of the proposed federated learning method - FedUCC.  
The overall architecture is composed of a central server and a number of clients, each of which has a multi-camera ReID dataset captured at different geographical locations. 
The task of the central server is to aggregate the trained models from these local clients, and redistribute an updated global model. 
The clients, on the other hand, are responsible for inheriting the global model's generic knowledge and conducting local optimization to properly integrate their specialised local knowledge.

The learning of FedUCC consists of three stages $s=\text{I, II, III}$ following a coarse-to-fine manner. Each training stage contains $T_s$ global round, including $E_s$ parallel client-wise (local) epochs and a global aggregation/redistribution. Particularly, the three stages are: 
I) \textbf{Generic knowledge guided modelling} formulates a model by adopting all available knowledge from the clients. II) \textbf{Specialized knowledge guided modelling} decouples the client-specific knowledge from the generic knowledge via parameter localization, and achieves personalized client models; and III) \textbf{Patch knowledge guided modelling} strengths the fine-grained patterns via local feature alignment.
The details of the proposed method are discussed in the rest of this section.

\subsection{Stage I: Generic Knowledge Guided Modelling}
At this stage, the goal is to obtain a global model that can roughly discriminate images across all the clients, of which statistical heterogeneity is not explicitly explored. It helps obtain a robust initial weight for the model to conduct further training in the next two stages.

Given $K$ clients, where the $k$-th client has a Person ReID dataset ${X^k}=\left\{x^{k}_{1}, x^k_{2}, \ldots, x^k_{n^k}\right\}$ with $n^k$ images, our
objective is to obtain model $\theta(W_g, W_l)$ on $\mathcal{X} = \{X^1, ..., X^k\}$. Note that the model $\theta(W_g, W_l)$ is parameterized by $(W_g, W_l)$ using shared generic parameters $W_g$ and client-specific  parameters $W_l$. 
Particularly, the specialized parameters $W_l$ are the neuron weights of batch normalization layers and the generic parameters are the remaining model parameters. We define ($W_g^k$ ,$W_l^k$) to be the trained generic parameters and specialized parameters on the $k$-th client, respectively.
At the current stage we focus on learning robust generic knowledge shared among clients, so both $(W_g, W_l)$ are updated to central server for redistribution. 
In the following part, we introduce the training steps taken by clients and the server respectively.

\subsubsection{Client Design}

Person ReID is a metric learning problem, thus at inference, the representations of image pairs capturing the same person are
supposed to be as close as possible. 
Therefore, a clustering algorithm is adopted, namely DBSCAN, where images with distance below a predefined threshold are grouped into the same cluster. 

Denote $\mathcal{H}^{k} = \{H^{k,1}, H^{k,2}, \dots, H^{k,m^k}\}$
as the clusters obtained from the $k$-th client with dataset $X^k$, where $m^k$ denotes the number of clusters in $X^k$. Note that each cluster $H^{k,i}$ is expected to contain all image features of a particular person. 
We first obtain a set of cluster centroids
$\mathcal{C}^k = \{c^{k}_1, \dots, c^{k}_{m^k}\}$ from $X^k$. 
Each cluster centroid is obtained by the mean feature vector of all images with the same pseudo-label:
\begin{equation}
\label{c_agg}
c^{k}_{j}=\frac{1}{\left|{H}^{k,j}\right|} \sum_{u^{k}_i \in {H}^{k,j}} u^{k}_i,
\end{equation}
where $u^{k}_i$ denotes the feature representation of image $x^{k}_i$, obtained from the model $\theta(W_g,W_l)$,  and 
 $\left|{H}^{k,j}\right|$ indicates the number of images in the cluster.

Cluster centroid contrastive learning focuses on cluster-level feature alignment. For each client $k$, the objective is to minimise the distance of each image feature to their cluster centroid feature via an InfoNCE loss~\cite{van2018representation}:
\begin{equation}
\label{l_info}
\mathcal{L}_{\text {u}}=\mathbb{E}\left[-\log \frac{\exp \left(u^k_i \cdot c^{k}_j / \tau\right)}{\sum_{j^\prime=1}^{m^k} \exp \left(u^k_i \cdot c^{k}_{^\prime} / \tau\right)}\right],
\end{equation}
where the distances between the query image feature $u^k_i$ and its cluster centroid $c^{k}_j$ are minimised and out-cluster centroid distances are maximised, optimizing with a similar principle as a softmax entropy. And $\tau$ is a temperature hyper-parameter.

During optimization, the centroid feature vectors $c^k_j$ of query instance $u^k_i$ are momentously updated in the memory dictionary as follows:
\begin{equation}
\label{momem}
c^{k}_j \leftarrow \lambda c^{k}_j+(1-\lambda)u^k_i,
\end{equation}
where $\lambda\in[0,1]$ is a hyper-parameter controlling the extent to update the cluster centroid. 
At the end of local client training, all parameters $(W^g, W^l)$ are forwarded to the server for multi-model aggregation. 

\subsubsection{Server Design}
The responsibility of the central server is to coordinate client knowledge sharing via an aggregation step. Basically, a strategy similar to FedAvg~\cite{mcmahan2017communication} is adopted in this study, which conducts the aggregation at the $(t+1)$-th global round as follows:
\begin{equation}
\label{mod_agg}
(W_g,W_l)^{t+1}=\sum_{k=1}^{K} \frac{n^{k}}{n} (W_g^k, W_l^k)^t,
\end{equation}
where $
n=\sum_{k=1}^{K} n^{k}$ are the total number of images of all clients. 
Instead of using equivalent average, the weighted sum is determined by the size of each client dataset. As a result, it allows the larger dataset clients to contribute richer knowledge for model aggregation, which is expected to contain more robust Person ReID patterns.

However, the issue of statistical heterogeneity is yet to be addressed. Existing methods attempted to alleviate this by combining the latest generic parameters with old specialized parameters to obtain a client-specific model~\cite{wu2021decentralised}. However, this doesn't fully resolve the distribution difference between clients as the model performance degrades after global aggregation/redistribution is still severe. Therefore, we further introduce the next modelling stage to solve this issue. 

\subsection{Stage II: Specialized Knowledge Guided Modelling}

The first stage has produced a generic model by taking into account the knowledge of all clients. This process aggregates all client-specific domain knowledge into the global model.
Due to the variety of different client data distributions, there are inevitably parameter collisions between the local models. The knowledge aggregation effect could be degraded, hence cause decreased performance on particular clients. 
We propose a knowledge guided modelling stage to further improve model accuracy by considering statistical heterogeneity.
It focuses on retrieving client-specific patterns to conduct knowledge personalization, which aims to maximize the model performance on all clients.



This stage is based on the local batch normalization (BN). 
Specifically, batch normalization layers are added in the backbone networks to normalize the activation. They adopt statistics in a mini-batch, which inherently captures the information related to the client data distribution~\cite{li2021fedbn}.
To achieve knowledge personalization, 
we preserve the learned local BN parameters by decoupling them from the global aggregation. This not only avoids parameter collision of the BN weights between clients but also allows the updated global model to be used in conjunction with the local BN parameters to achieve personalization.
At this stage, the model aggregation and updates are as follows: 
\begin{equation}
\label{mod_agg_gl}
(W_g)^{t+1}=\sum_{k=1}^{K} \frac{n^{k}}{n} (W_g^k)^{t}.
\end{equation}
And model localization for the $k$-th client is as follows:
\begin{equation}
\label{mod_agg_gl}
\theta^{k,t+1}=\theta((W_g)^{t+1}, (W_l^k)^{t}).
\end{equation}


Note that BN layers for personalization could lead to sub-optimal performance due to the over-fitting issue on a client. We further enhance image feature learning by introducing patch-level representation to enable
fine-grained aspects for facilitating parameter alignment across clients.









\subsection{Stage III: Patch Knowledge Guided Modelling}
Patch-based representations show their advantages for the robustness on unseen identities which are not involved in the training process~\cite{yang2019patch}. 
Hence, additional model parameters which are used to characterize patch-level features are introduced and optimized by following a similar strategy as in Stage II. 

Specifically, given the dataset $X^k$ on the $k$-th client, we obtain the set of 3D tensor features of images before the global average pooling (GAP) layer in the backbone model as $V^k = \{v^k_1, \dots, v^k_{n^k}\}$.
We then extract the set of patch features for all images as $\mathcal{Q}^{k} = \{Q^{k}_{1}, \dots, Q^{k}_{n^k}\}$
by equally splitting $v^k_i$ into $p$ horizontal strips and further adopt GAP to obtain patch-level image features. In detail, we represent them as $Q^{k}_{i} = \{q^{k}_{i,1}, \dots, q^{k}_{i,p}\}$.

Finally, for each cluster $H^{k,j} \in \mathcal{H}^k$, we further extend it to $p$ clusters for features within particular patches, of which the centroids are: 
$D^{k}_j = \{d^k_{j,1}, \dots, d^k_{j,p}\}$
associated with the corresponding patch-level features. 


After that, given a patch feature representation ${q}^k_{i,r}$ and a centroid ${d}^k_{j,s}$, a loss function can be defined as follows:
\begin{equation}
\label{lp_info}
\mathcal{L}_{\text {p}}=\mathbb{E}\sum_{r=1}^{p}{\left[-\log \frac{\exp \left(q^{k}_{i,r} \cdot d^{k}_{j,s} / \tau\right)}{\sum_{j^\prime=1}^{m^K} \exp \left(q^{k}_{i,r} \cdot d^{k}_{j^\prime,s} / \tau\right)}\right]},
\end{equation}
which is similar to Eq.(\ref{l_info}).

To this end, a loss function jointly optimized $L_u$ and $L_{p}$ can be adopted to achieve a robust federated Person ReID:
\begin{equation}
\label{lp_al}
\mathcal{L} = \mathcal{L}_{u} + \mathcal{L}_{p}.
\end{equation}

\begin{table*}[!htbp]
\begin{adjustwidth}{-0.35in}{-0.35in}
\caption{\label{Tab:1}Overall performance with ablation studies on eight Person ReID datasets.}
\small
\begin{adjustbox}{width=0.96\textwidth,center}
\begin{tabular}{   p{2.0 cm} |  * {8}  { * 2 { |p{0.5cm}}|}| p{0.5 cm} | p{0.5 cm} }
\hline
Methods & 
\multicolumn{2}{c||}{Duke} &
\multicolumn{2}{c||}{Market}  &
\multicolumn{2}{c||}{CUHK03} &
\multicolumn{2}{c||}{PRID} &
\multicolumn{2}{c||}{CUHK01}  &
\multicolumn{2}{c||}{VIPeR}& 
\multicolumn{2}{c||}{3Dpes} &
\multicolumn{2}{c||}{iLIDS} & 
\multicolumn{2}{c} {Avg}\\
\hline

Metric (\%) & R1 &  mAP  & R1 &  mAP & R1 &  mAP & R1 &  mAP & R1 &  mAP & R1 &  mAP & R1 &  mAP & R1 &  mAP & R1 & mAP \\
\hline

FedUReID~\cite{zhuang2021joint} & 51.0 & - & 65.2 & -& 8.9& - & 38.0& - & 43.6& -& 26.6 & -& 65.3&- &73.5 &- & 46.5&-\\
\hline

Backbone & 
14.4&6.6&42.2&21.3&1.4&2.0&19.9&24.4&26.7&26.1&22.7&28.2&61.2&43.4&63.9&48.5& 31.5 & 25.1\\
Stage I & 
65.7& 44.4 &
74.6 & 45.6 &
\textbf{11.3} & 9.4 & 
24.0 & 27.2 & 
60.6 & 57.3 &
\textbf{35.4} & \textbf{39.7} &
60.3 & 41.7 &
\textbf{81.0} & \textbf{66.2} &
51.6& 41.4\\
Stage I + II & 75.2&55.6&85.7&63.5&8.9&9.0&52.9&57.2&69.5&65.7&31.0&36.1&65.5&46.4&80.1&60.7&58.6&49.3\\
Stage I + III & 76.3&56.3&82.6&58.2&9.6&8.8&46.9&52.1&74.5&71.4&28.1&33.1&65.5&46.6&67.5&53.6&56.4&47.5\\
FedUCC & \textbf{78.8}&\textbf{60.5}&\textbf{86.5}&\textbf{65.5}&9.6&\textbf{9.7}&\textbf{58.9}&\textbf{63.1}&\textbf{78.3}&\textbf{75.3}&31.3&36.7&\textbf{68.9}&\textbf{50.9}&74.7&59.7&\textbf{60.9}&\textbf{52.7}\\
\hline

\end{tabular}
\end{adjustbox}
\label{tab:cross}
\end{adjustwidth}

\end{table*}

\begin{table*}[!htbp]
\centering
\caption{\label{Tab:2}Comparisons to the state-of-the-art methods.}

\begin{tabular}{  | p{3 cm}|| p{3.5 cm} | * {2}  { * 4 { |p{0.8cm}}|} }
\hline
Methods  & Types &  \multicolumn{4}{c||}{Duke} & \multicolumn{4}{c|}{Market} \\
\hline

Metric (\%)  && R1 &  R5 & R10 &  mAP & R1 &  R5 & R10 &  mAP  \\
\hline

PUL~\cite{qiao2018few} &Domain Adaptation& 
30.4& 46.4& 50.7 &16.4 & 
44.7& 59.1 &65.6 &20.1 
\\
HHL~\cite{zhong2018generalizing}&Domain Adaptation& 
46.9& 61.0 &66.7& 27.2 &
62.2& 78.8& 84.0& 31.4 
\\
SPGAN~\cite{deng2018image} & Domain Adaptation& 
46.9& 62.6& 68.5 &26.4 &
58.1& 76.0 &82.7& 26.7
\\
BUC~\cite{lin2019bottom}& Purely Unsupervised & 40.4 & 52.5 & 58.2 & 22.1 & 61.9 & 73.5 & 78.2 & 29.6 \\

MMCL~\cite{wang2020unsupervised} & Purely Unsupervised &  65.2 &75.9 &80.0& 40.2 & 
80.3 &89.4& 92.3& 45.5  \\
FedUReID~\cite{zhuang2021joint} & Federated Unsupervised & 
51.0& 62.4 &67.6 &29.5 & 
65.2& 77.8 &82.2 &34.2 
\\

\hline
FedUCC (Ours) &Federated Unsupervised 
&\textbf{78.8}&\textbf{87.2}&\textbf{89.8}&\textbf{60.5}&\textbf{86.5}&\textbf{94.5}&\textbf{96.7}&\textbf{65.5}\\


\hline
\end{tabular}
\label{tab:cross}
\end{table*}

\section{Experiments}
\subsection{Dataset}
Seven Person Re-ID datasets were adopted to comprehensively evaluate our proposed method, including DukeMTMC-ReID~\cite{ristani2016performance},
Market1501~\cite{zheng2015scalable},  CUHK03-NP~\cite{li2014deepreid},
PRID~\cite{hirzer2011person},
CUHK01~\cite{li2012human},
VIPeR~\cite{gray2008viewpoint},
3DPeS~\cite{Baltieri20113DPeS3P} and
iLIDS~\cite{wang2014person}. 
Based on the federated learning setting, each dataset is considered to be held by a single client. Therefore, each client's data is collected at a unique location. This is consistent to real-world scenarios when delivering a Person Re-ID model with the consideration of the privacy requirements. 
 
Standard Person ReID evaluation metrics were adopted to measure the performance, including Cumulative Match Characteristic (CMC) curve and mean Average Precision (mAP)~\cite{zheng2016person}. 
Given a query image and the gallery containing a group of watch list images, CMC first ranks the gallery images in line with their similarity to the query. Then, each image ranked within the rank-K (i.e., top-K) results is verified whether a matching identity is within the results or not. 
In addition, the rank-1 accuracy is introduced to indicate the average probability of the most similar image matching the query image, and mAP reflects the mean average precision of all queries.

\subsection{Experimental Settings}
\textbf{Backbone Architectures} ResNet-50~\cite{he2016deep}  pretrained on ImagNet~\cite{deng2009imagenet} was adopted as the backbone network for feature extraction.

\textbf{Implementation Details}
For each client, an Adam optimizer with Nesterov momentum 0.9 
was adopted. The learning rate was set to $3.5\times 10^{-4}$ for local model optimization, and the batch size was set to 64 empirically. 
The number of patch-feature segmentation  $N_p$ was set to 2, indicating the upper-half and lower-half of the human body.
We employed DBSCAN~\cite{ester1996density} for image feature clustering based on Jaccard distance with k-reciprocal encoding~\cite{zhong2017re} and set the number of minimum samples as 2 to discard un-clustered images.
The implementation was with an NVIDIA V100 for training and inference stages. 

\textbf{Training Settings}
The model parameters were optimized through 3 training stages as discussed in Sec. \ref{sec:method}. 
The numbers of local epoch and global round differ between the 3 training stages. Stage I training was optimized with local epoch $E_I =  1$ and global round $T_I = 100$, which rapidly aggregated and updated the model parameters from the local clients. 
For optimization at Stage II and Stage III, we empirically set the local epoch $E_{II} = E_{III} = 5$ and global round $T_{II} = T_{III} = 100$, allowing Stage II and Stage III to learn sufficient local knowledge. 
In addition, we concatenated the features extracted from a part-based model and from the Stage II model for the clustering and training of Stage III.




\begin{figure}[!tbp]
  \centering
  \includegraphics[width=0.31\textwidth]{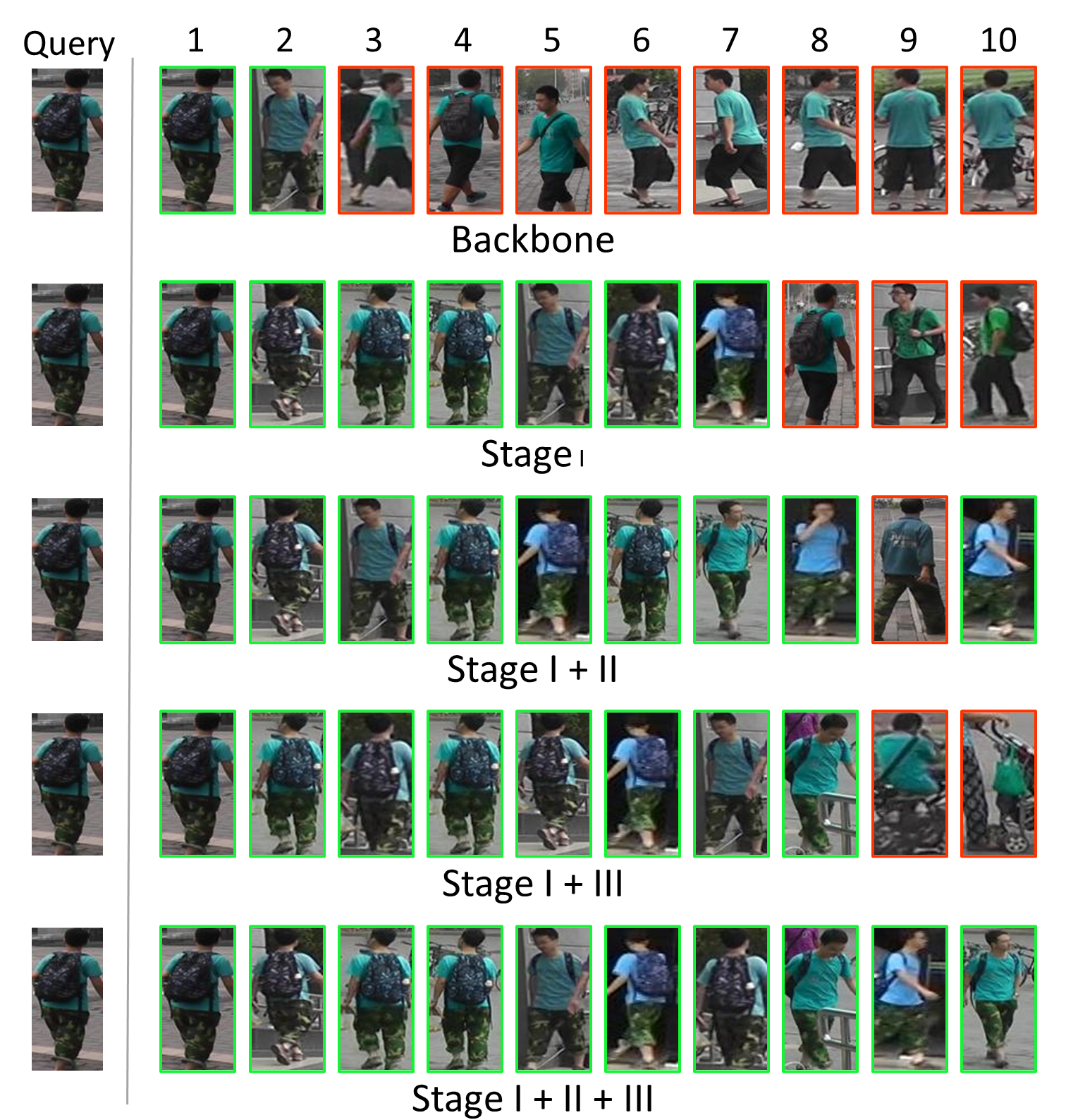}\label{fig:r1}
  \caption{Ranking result on three queries}
  \label{fig:ranking}
\end{figure}

\subsection{Comparisons with the State-of-the-Art  Unsupervised ReID}

We compared the proposed method with a 
number of existing methods following their evaluation protocols 
to demonstrate the effectiveness of our method. 
In Table~\ref{Tab:1}, 
a recently proposed unsupervised federated Person Re-ID method FedUReID~\cite{wu2021decentralised} was compared to ours on 8 datasets. 
Our proposed method performed consistently better than FedUReID on all datasets. 
In particular, our result outperforms FedUReID with a large margin on the Duke, Market, and CUHK01 datasets.
Note that Duke and Market are the two largest datasets. Such result indicates significant incremental in terms of the top-1 accuracy from 51.0\% to 77.8\% on Duke and 65.2\% to 86.8\% on Market. 
Other datasets also demonstrate the effectiveness of FedUCC. CUHK01 shows the largest improvement regarding the top-1 accuracy 43.6\% to 78.3\%, PRID shows 20.9\% incremental, and the remaining CUHK03, VIPeR, 3DPeS, iLIDS also show perofrmance improvement of 0.7\%, 4.7\%, 3.6\% and 1.2\%, respectively. 

In Table~\ref{Tab:2}, FedUCC is further compared with two additional categories of methods including the unsupervised domain adaption (PUL~\cite{qiao2018few},
HHL~\cite{zhong2018generalizing}, and
SPGAN~\cite{deng2018image}) and the fully unsupervised approaches (BUC~\cite{lin2019bottom} and MMCL~\cite{wang2020unsupervised}). 
The evaluation was conducted on the two largest datasets: Duke and Market. Note that the unsupervised domain adaptation requires the source dataset to have annotated labels. 
Without the additional supervision using annotations, our method still performs better than these domain adaptation methods. For example, our method and SPGAN~\cite{deng2018image} achieve 77.8\% and 46.9\% rank-1 accuracy on Duke, respectively; our method and HHL~\cite{zhong2018generalizing} achieve 86.8\% and 62.2\% rank-1 accuracy on Market. Additionally, we compare our results with fully unsupervised methods, where each ReID model is trained independently on individual datasets without collaboration. Superior performance was also consistently demonstrated, given the comparison of our result to MMCL~\cite{wang2020unsupervised} with 77.8\% and 65.2\% rank-1 accuracy on Duke,  86.6\% and 80.3\% on Market. This also highlights the advantage of collaboration training using federated learning.



\begin{figure}[t]
\begin{center}
  \includegraphics[width=1\linewidth]{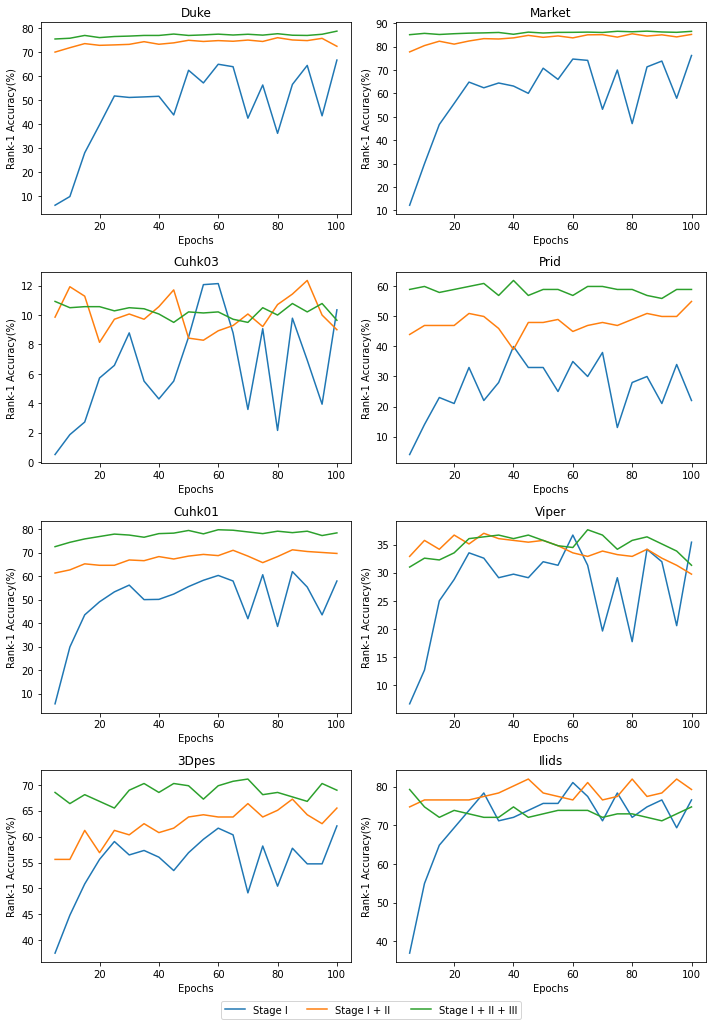}
\end{center}
  \caption{
Rank-1 accuracy curves on all datasets
}
\label{fig:r1-accs}
\end{figure}


\subsection{Ablation Studies}

Ablation studies were conducted to demonstrate the effectiveness of the individual strategies proposed, which helped explore the role of each stage in the entire pipeline of FedUCC.  
Particularly, Table~\ref{Tab:1} lists the comparisons of several architectures in terms of Rank-1 accuracy and mAP. These architectures include a backbone only method, a Stage I only method, Stage I + II method and Stage I + III method. 
Overall, 
our proposed FedUCC achieves the best overall performance with all these strategies. We discuss the details of these individual architectures in the following sections.

\textbf{Backbone Method: Generic Knowledge Guided Modelling} 
The backbone setting adopted conventional DBSCAN-based label assignment and conducted optimization using a cross-entropy loss for classification. 
For federated learning, we chose a local epoch of 5 that balances the model convergence and the cross-client knowledge exchange. 
Each client model was updated with a fixed number of training batches in line with the dataset size. FedAvg considered this by limiting smaller client models with fewer averaging ratios for aggregation. 
However, skewness can usually be identified among the clients and the local model parameters were likely to conflict.
As a result, the local training led to significant fluctuation of the global model performance between the global rounds and eventually limited the person ReID accuracy. 

\textbf{Stage I: Generic Knowledge Guided Modelling with Memory-based Contrastive Learning}
Compared to the backbone method, Stage I further introduces the memory-based contrastive learning on each client as indicated in Eq. (\ref{c_agg}) - (\ref{momem}). 
As shown in TABLE~\ref{Tab:1}, the performance was superior on all clients. Particularly, the Rank-1 accuracy increased on the largest dataset Duke and Market 51.3\% and 32.4\% and achieved an average increase of 20.1\% on all datasets. 
Such performance improvement shows the reduced influence of statistical heterogeneity. Due to the more effective model training strategy and the unresolved issue of client knowledge conflict, the model performance fluctuation on all clients is still severe as illustrated in Fig.~\ref{fig:r1-accs}.

\textbf{Stage I + II: Specialized Knowledge Guided Modelling with Batch Normalization} 
Stage II helps explore personalized knowledge, which retained the batch normalization patterns within each client. 
Specifically, Stage II has shown an average of 7.9\% increase in terms of the Rank-1 accuracy.
Additionally, the model performance is much more stable compared with Stage I,
as shown in Fig.~\ref{fig:r1-accs}.
This indicates the problem of statistical heterogeneity is greatly reduced with effective personal knowledge decoupling.
Note that the improvements were not consistent across all the clients. For smaller datasets such as CUHK03, VIPeR, and iLIDS, the performance even dropped slightly. 
It suggested a potential domain over-fitting issue, such that the generic knowledge from the shared parameters was overwhelmed by the local knowledge. Therefore, this motivated us to further explore patch-based feature learning to enhance the robustness of the feature spaces.

\textbf{Stage I+III: Patch Knowledge Guided Modelling}
The patch-based features were explored for fine-grained person representation.
As shown in TABLE~\ref{Tab:1}, the average Rank-1 accuracy has decreased by 1.8\% on average compared with using holistic person feature. However by ensemble both holistic and patch features (FedUCC), better discriminability was achieved compared with using standalone representation, with an average increase of rank-1 accuracy by 3.4\% 
Specifically, consistent performance gain can be observed on the majority of those datasets 
(i.e., CUHK03, PRID, CUHK01, VIPeR, 3DPeS, iLIDS) compared with the strategy of Stage I + II. 
In our observations, iLIDS shows the highest performance in Stage I training, and no performance gain has been shown for later stages of training.
Given that it is the smallest dataset, the number of images is likely insufficient for the Stage I model to converge better.


\subsection{Qualitative Study}
Fig.~\ref{fig:ranking} visualizes the top-10 results of three queries to demonstrate the effectiveness of our three-stage modelling method.
It can be seen that the backbone-only method retrieves the most incorrect images. Even though the general appearances are similar, attention to more detailed patterns is lacking.
Visual clues such as the backpack are rarely matched in the corresponding query results.
Stage I shows a much better retrieval accuracy in Fig.~\ref{fig:ranking}, where more discriminative patterns are identified to mine the matching query images. 
Stage II training further improves retrieval accuracy with enhanced ability to mine discriminative patterns despite viewpoint variations.
Better retrieval accuracy are shown with more front and side views of the person despite the larger visual variation introduced by the backpack. 
Stage III model shows a comparable result to Stage II, with more focus on regional pattern matching. This is indicated by the 10-th negative retrieval image where the camo patterns in the query's backpack and short are closely matched.
Lastly, the FedUCC (Stage I + II + III) model ensembles holistic and patch-level features to obtain a refined feature representation, improving query results with both holistic and fine-grained patterns, as illustrated in Fig.~\ref{fig:ranking}.

\section{Conclusion}
In this paper, we present a federated unsupervised cluster-contrastive learning method, namely FedUCC, to address the Person ReID problem. FedUCC introduces a three-stage modelling strategy following a coarse-to-fine manner. In detail, generic knowledge, specialized knowledge and patch knowledge are explored with a deep neural network. It enables the sharing of mutual knowledge while retaining local domain-specified knowledge through network layers and their parameters.
Comprehensive experiments on 8 public benchmark datasets consistently demonstrated the perofrmance improvement of our proposed FedUCC, compared with existing methods.
For our future work, we will consider the exploitation of outlier images. As during clustering, a small number of images in a cluster are considered outliers and are discarded to reduce their negative impacts. However, those are still important assets with the potential to improve the capability with better feature diversity.






\bibliographystyle{IEEEtran}

\bibliography{reference}

\end{document}